\documentclass[10pt,twocolumn,letterpaper]{article}
\usepackage{iccv}
\usepackage{times}
\usepackage{epsfig}
\usepackage{graphicx}
\usepackage{amsmath}
\usepackage{amssymb}
\usepackage{caption}
\usepackage{subcaption}
\usepackage{mathptmx} 
\usepackage{color}
\usepackage{colortbl}
\usepackage{enumitem}
\usepackage{booktabs}
\usepackage{tablefootnote}
\usepackage[toc,page]{appendix}

\usepackage[pagebackref=true,breaklinks=true,letterpaper=true,colorlinks,bookmarks=false]{hyperref}

\iccvfinalcopy

\ificcvfinal\fi
\begin{document}

\title{An Analysis of Visual Question Answering Algorithms}
\author{Kushal Kafle \qquad Christopher Kanan\thanks{Corresponding author}\\
Rochester Institute of Technology\\
Rochester, New York\\
{\tt\small kk6055,kanan@rit.edu}}
\maketitle
\begin{abstract}

In visual question answering (VQA), an algorithm must answer text-based questions about images. While multiple datasets for VQA have been created since late 2014, they all have flaws in both their content and the way algorithms are evaluated on them. As a result, evaluation scores are inflated and predominantly determined by answering easier questions, making it difficult to compare different methods. In this paper, we analyze existing VQA algorithms using a new dataset called the Task Driven Image Understanding Challenge (TDIUC), which has over 1.6 million questions organized into 12 different categories (available for download at \url{https://goo.gl/Ng9ix4}). We also introduce questions that are meaningless for a given image to force a VQA system to reason about image content. We propose new evaluation schemes that compensate for over-represented question-types and make it easier to study the strengths and weaknesses of algorithms. We analyze the performance of both baseline and state-of-the-art VQA models, including multi-modal compact bilinear pooling (MCB), neural module networks, and recurrent answering units. Our experiments establish how attention helps certain categories more than others,  determine which models work better than others, and explain how simple models (e.g. MLP) can surpass more complex models (MCB) by simply learning to answer large, easy question categories.
\end{abstract}

\section{Introduction}

In open-ended visual question answering (VQA) an algorithm must produce answers to arbitrary text-based questions about images~\cite{malinowski2014multi,antol2015vqa}. VQA is an exciting computer vision problem that requires a system to be capable of many tasks. Truly solving VQA would be a milestone in artificial intelligence, and would significantly advance human computer interaction. However, VQA datasets must test a wide range of abilities for progress to be adequately measured.

VQA research began in earnest in late 2014 when the DAQUAR dataset was released~\cite{malinowski2014multi}. Including DAQUAR, six major VQA datasets have been released, and algorithms have rapidly improved. 
On the most popular dataset, `The VQA Dataset'~\cite{antol2015vqa}, the best algorithms are now approaching 70\% accuracy~\cite{FukuiPYRDR16} (human performance is 83\%). While these results are promising, there are critical problems with existing datasets in terms of multiple kinds of biases. Moreover, because existing datasets do not group instances into meaningful categories, it is not easy to compare the abilities of individual algorithms. For example, one method may excel at color questions compared to answering questions requiring spatial reasoning. Because color questions are far more common in the dataset, an algorithm that performs well at spatial reasoning will not be appropriately rewarded for that feat due to the evaluation metrics that are used.

\begin{figure}
\centering
\includegraphics[width=0.45\textwidth]{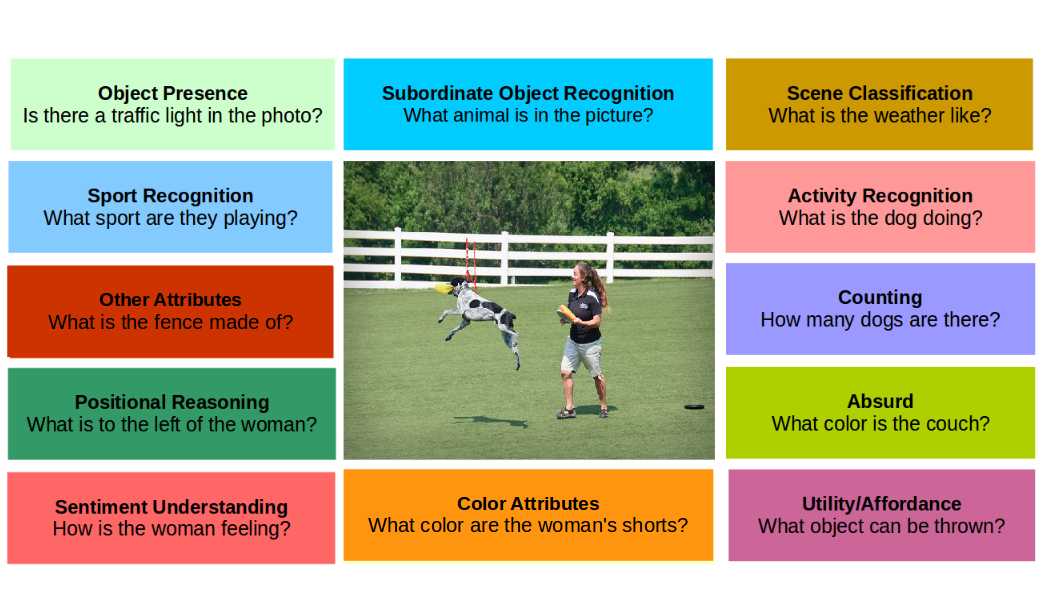}\\
\caption{A good VQA benchmark tests a wide range of computer vision tasks in an unbiased manner. In this paper, we propose a new dataset with 12 distinct tasks and evaluation metrics that compensate for bias, so that the strengths and limitations of  algorithms can be better measured.}
\label{tbl:overview} 
\end{figure}

\textbf{Contributions}: Our paper has four major contributions aimed at better analyzing and comparing VQA algorithms: 1) We create a new VQA benchmark dataset where questions are divided into 12 different categories based on the task they solve; 2) We propose two new evaluation metrics that compensate for forms of dataset bias; 3) We balance the number of yes/no object presence detection questions to assess whether a balanced distribution can help algorithms learn better; and 4) We introduce absurd questions that force an algorithm to determine if a question is valid for a given image. We then use the new dataset to re-train and evaluate both baseline and state-of-the-art VQA algorithms. We found that our proposed approach enables more nuanced comparisons of VQA algorithms, and helps us understand the benefits of specific techniques better. In addition, it also allowed us to answer several key questions about VQA algorithms, such as, `Is the generalization capacity of the algorithms hindered by the bias in the dataset?', `Does the use of spatial attention help answer specific question-types?', `How successful are the VQA algorithms in answering less-common questions?', and 'Can the VQA algorithms differentiate between real and absurd questions?'


\section{Background} 
\subsection{Prior Natural Image VQA Datasets}
Six datasets for VQA with natural images have been released between 2014--2016: DAQUAR~\cite{malinowski2014multi}, COCO-QA~\cite{ren2015image}, FM-IQA~\cite{gao2015you}, The VQA Dataset~\cite{antol2015vqa}, Visual7W~\cite{visual7w}, and Visual Genome~\cite{krishnavisualgenome}. FM-IQA needs human judges and has not been widely used, so we do not discuss it further. Table~\ref{tab:datasets} shows statistics for the other datasets. Following others~\cite{Kafle2016,zhou2015,WuWSHD15}, we refer to the portion of The VQA Dataset containing natural images as COCO-VQA. Detailed dataset reviews can be found in \cite{kafle2016review} and \cite{wu2016visual}.

All of the aforementioned VQA datasets are biased. DAQUAR and COCO-QA are small and have a limited variety of question-types. Visual Genome, Visual7W, and COCO-VQA are larger, but they suffer from several biases. Bias takes the form of both the kinds of questions asked and the answers that people give for them. For COCO-VQA, a system trained using only question features achieves 50\% accuracy~\cite{Kafle2016}. This suggests that some questions have predictable answers.  Without a more nuanced analysis, it is challenging to determine what kinds of questions are more dependent on the image. For datasets made using Mechanical Turk, annotators often ask object recognition questions, e.g., `What is in the image?' or `Is there an elephant in the image?'. Note that in the latter example, annotators rarely ask that kind of question unless the object is in the image. On COCO-VQA, 79\% of questions beginning with `Is there a' will have `yes' as their ground truth answer.

In 2017, the VQA 2.0~\cite{balanced_vqa_v2}  dataset was introduced. In VQA 2.0, the same question is asked for two different images and annotators are instructed to give opposite answers, which helped reduce language bias. However, in addition to language bias, these datasets are also biased in their distribution of different types of questions and the distribution of answers within each question-type. Existing VQA datasets use performance metrics that treat each test instance with equal value (e.g., simple accuracy). While some do compute additional statistics for basic question-types, overall performance is not computed from these sub-scores~\cite{antol2015vqa,ren2015image}. This exacerbates the issues with the bias because the question-types that are more likely to be biased are also more common. Questions beginning with `Why' and `Where' are rarely asked by annotators compared to those beginning with `Is' and 'Are'. For example, on COCO-VQA, improving accuracy on `Is/Are' questions by 15\% will increase overall accuracy by over 5\%, but answering \emph{all} `Why/Where' questions correctly will increase accuracy by only 4.1\%~\cite{kafle2016review}. Due to the inability of the existing evaluation metrics to properly address these biases, algorithms trained on these datasets learn to exploit these biases, resulting in systems that work poorly when deployed in the real-world. 

For related reasons, major benchmarks released in the last decade do not use simple accuracy for evaluating image recognition and related computer vision tasks, but instead use metrics such as mean-per-class accuracy that compensates for unbalanced categories. For example, on Caltech-101~\cite{Fei-Fei2006}, even with balanced training data, simple accuracy fails to address the fact that some categories were much easier to classify than others (e.g., faces and planes were easy and also had the largest number of test images). Mean per-class accuracy compensates for this by requiring a system to do well on each category, even when the amount of test instances in categories vary considerably.

Existing benchmarks do not require reporting accuracies across different question-types. Even when they are reported, the question-types can be too coarse to be useful, e.g., `yes/no', `number' and `other' in COCO-VQA. To improve the analysis of the VQA algorithms, we categorize the questions into meaningful types, calculate the sub-scores, and incorporate them in our evaluation metrics.

\subsection{Synthetic Datasets that Fight Bias}
Previous works have studied bias in VQA and proposed countermeasures. In \cite{Zhang_2016_CVPR}, the Yin and Yang dataset was created to study the effect of having an equal number of binary (yes/no) questions about cartoon images. They found that answering questions from a balanced dataset was harder. This work is significant, but it was limited to yes/no questions and their approach using cartoon imagery cannot be directly extended to real-world images.

One of the goals of this paper is to determine what kinds of questions an algorithm can answer easily. In \cite{AndreasRDK15}, the SHAPES dataset was proposed, which has similar objectives. SHAPES is a small dataset, consisting of 64 images that are composed by arranging colored geometric shapes in different spatial orientations. Each image has the same 244 yes/no questions, resulting in 15,616 questions. Although SHAPES serves as an important adjunct evaluation, it alone cannot suffice for testing a VQA algorithm. The major limitation of SHAPES is that all of its images are of 2D shapes, which are not representative of real-world imagery. Along similar lines, Compositional Language and Elementary Visual Reasoning (CLEVR)~\cite{johnson2016clevr} also proposes use of 3D rendered geometric objects to study reasoning capacities of a model. CLEVR is larger than SHAPES and makes use of 3D rendered geometric objects. In addition to shape and color, it adds material property to the objects. CLEVR has five types of questions: attribute query, attribute comparison, integer comparison, counting, and existence.

Both SHAPES and CLEVR were specifically tailored for compositional language approaches~\cite{AndreasRDK15} and downplay the importance of visual reasoning. For instance, the CLEVR question, `What size is the cylinder that is left of the brown metal thing that is left of the big sphere?' requires demanding language reasoning capabilities, but only limited visual understanding is needed to parse simple geometric objects. Unlike these three synthetic datasets, our dataset contains natural images and questions. To improve algorithm analysis and comparison, our dataset has more (12) explicitly defined question-types and new evaluation metrics.

\begin{table*}[t]
\centering
\footnotesize
\caption{Comparison of previous natural image VQA datasets with TDIUC. For COCO-VQA, the explicitly defined number of question-types is used, but a much finer granularity would be possible if they were individually classified. MC/OE refers to whether open-ended or multiple-choice evaluation is used.}
\label{tab:datasets}
\begin{tabular}{lrrccrc}
\toprule
                            & \multicolumn{1}{c}{\textbf{\begin{tabular}[c]{@{}c@{}} Images\end{tabular}}} & \multicolumn{1}{c}{\textbf{Questions}} & \textbf{\begin{tabular}[c]{@{}c@{}}Annotation\\ Source\end{tabular}} & \textbf{\begin{tabular}[c]{@{}c@{}}Question\\ Types\end{tabular}} & \multicolumn{1}{c}{\textbf{\begin{tabular}[c]{@{}c@{}}Unique\\ Answers\end{tabular}}} & \textbf{MC/OE} \\ \midrule 
\textbf{DAQUAR}             & 1,449             & 16,590       & Both   	  & 3        & 968       		& OE             \\
\textbf{COCO-QA}            & 123,287           & 117,684      & Auto     	  & 4        & 430       		& OE             \\
\textbf{COCO-VQA}           & 204,721           & 614,163      & Manual   	  & 3        & 145,172          & Both           \\
\textbf{Visual7W}          & 47,300             & 327,939      & Manual       & 7        & 25,553           & MC             \\
\textbf{Visual Genome}      & 108,000           & 1,773,358    & Manual       & 6        & 207,675          & OE             \\
\textbf{TDIUC (This Paper)} & 167,437           & 1,654,167    & Both         & 12       & 1,618            & OE  \\  
\bottomrule
\end{tabular}
\end{table*}

\section{TDIUC for Nuanced VQA Analysis}

\begin{figure}[t]
\footnotesize
	\centering
    \captionsetup[subfigure]{labelformat=empty}    
        \begin{subfigure}[t]{0.23\textwidth}
		\includegraphics[width=\textwidth,height=3.5cm]{images/fig1.jpg}
		\caption{\footnotesize \textbf{Q:} What color is the suitcase? \textbf{A:} Absurd \textbf{Q:} What color is the man's hat? \textbf{A: }White \textbf{Q:} What sport is this?  \textbf{A:} Tennis}%
	\end{subfigure}
    \hfill
    \begin{subfigure}[t]{0.23\textwidth}
	 	\includegraphics[width=\textwidth,height=3.5cm]{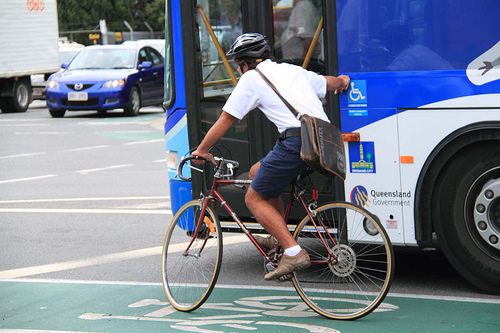}
		\caption{\footnotesize \textbf{Q:} What is to the left of the blue bus?  \textbf{A:} Car \textbf{Q:} Is there a train in the photo?  \textbf{A:} No \textbf{Q:} How many bicycles are there?  \textbf{A:} One}%
	\end{subfigure}
        \caption{Images from TDIUC and their corresponding question-answer pairs.}
    \label{fig:dataset-examples}
\end{figure}

In the past two years, multiple publicly released datasets have spurred the VQA research. However, due to the biases and issues with evaluation metrics, interpreting and comparing the performance of VQA systems can be opaque. We propose a new benchmark dataset that explicitly assigns questions into 12 distinct categories. This enables measuring performance within each category and understand which kind of questions are easy or hard for today's best systems. Additionally, we use evaluation metrics that further compensate for the biases. We call the dataset the Task Driven Image Understanding Challenge (TDIUC). The overall statistics and example images of this dataset are shown in 
Table~\ref{tab:datasets} and Fig.~\ref{fig:dataset-examples} respectively.

TDIUC has 12 question-types that were chosen to represent both classical computer vision tasks and novel high-level vision tasks which require varying degrees of image understanding and reasoning. The question-types are: 
\begin{enumerate}[noitemsep]
\item Object Presence (e.g., `Is there a cat in the image?')
\item Subordinate Object Recognition (e.g., `What kind of furniture is in the picture?')
\item Counting (e.g., 'How many horses are there?')
\item Color Attributes (e.g., `What color is the man's tie?')
\item Other Attributes (e.g., `What shape is the clock?')
\item Activity Recognition (e.g., `What is the girl doing?') 
\item Sport Recognition (e.g.,`What are they playing?')
\item Positional Reasoning (e.g., `What is to the left of the man on the sofa?')
\item Scene Classification (e.g., `What room is this?')
\item Sentiment Understanding (e.g.,`How is she feeling?')
\item Object Utilities and Affordances (e.g.,`What object can be used to break glass?')
\item Absurd (i.e., Nonsensical queries about the image)
\end{enumerate}
The number of each question-type in TDIUC is given in Table~\ref{tab:num-types}. The questions come from three sources. First, we imported a subset of questions from COCO-VQA and Visual Genome. Second, we created algorithms that generated questions from COCO's semantic segmentation annotations~\cite{lin2014microsoft}, and Visual Genome's objects and attributes annotations~\cite{krishnavisualgenome}. Third, we used human annotators for certain question-types. In the following sections, we briefly describe each of these methods.

\subsection{Importing Questions from Existing Datasets}

We imported questions from COCO-VQA and Visual Genome belonging to all question-types except `object utilities and affordances'. We did this by using a large number of templates and regular expressions. For Visual Genome, we imported questions that had one word answers. For COCO-VQA, we imported questions with one or two word answers and in which five or more annotators agreed.

For color questions, a question would be imported if it contained the word `color' in it and the answer was a commonly used color. Questions were classified as activity or sports recognition questions if the answer was one of nine common sports or one of fifteen common activities and the question contained common verbs describing actions or sports, e.g., playing, throwing, etc. For counting, the question had to begin with `How many' and the answer had to be a small countable integer (1-16). The other categories were determined using regular expressions. For example, a question of the form `Are \line(1,0){15} feeling  \line(1,0){15}?' was classified as sentiment understanding and `What is to the right of/left of/ behind the \line(1,0){15}?' was classified as positional reasoning. Similarly, `What \texttt{<OBJECT CATEGORY>} is in the image?' and similar templates were used to populate subordinate object recognition questions. This method was used for questions about the season and weather as well, e.g., `What season is this?', `Is this rainy/sunny/cloudy?', or `What is the weather like?' were imported 
to scene classification.

\subsection{Generating Questions using Image Annotations}

Images in the COCO dataset and Visual Genome both have individual regions with semantic knowledge attached to them. We exploit this information to generate new questions using question templates. To introduce variety, we define multiple templates for each question-type and use the annotations to populate them. For example, for counting we use $8$ templates, e.g., `How many \texttt{<objects>} are there?',  `How many \texttt{<objects>} are in the photo?', etc. Since the COCO and Visual Genome use different annotation formats, we discuss them separately.

\subsubsection{Questions Using COCO annotations}

Sport recognition, counting, subordinate object recognition, object presence, scene understanding, positional reasoning, and absurd questions were created from COCO, similar to the scheme used in \cite{kafle2017b}. For \textbf{counting}, we count the number of object instances in an image annotation. To minimize ambiguity, this was only done if objects covered an area of at least 2,000 pixels.

For \textbf{subordinate object recognition}, we create questions that require identifying an object's subordinate-level object classification based on its larger semantic category. To do this, we use COCO supercategories, which are semantic concepts encompassing several objects under a common theme, e.g., the supercategory `furniture' contains chair, couch, etc. If the image contains only one type of furniture, then a question similar to `What kind of furniture is in the picture?' is generated because the answer is not ambiguous. Using similar heuristics, we create questions about identifying food, electronic appliances, kitchen appliances, animals, and vehicles. 

For \textbf{object presence} questions, we find images with objects that have an area larger than 2,000 pixels and produce a question similar to `Is there a \texttt{<object>} in the picture?' These questions will have `yes' as an answer. To create negative questions, we ask questions about COCO objects that are not present in an image. To make this harder, we prioritize the creation of questions referring to absent objects that belong to the same supercategory of objects that are present in the image. A street scene is more likely to contain trucks and cars than it is to contain couches and televisions. Therefore, it is more difficult to answer `Is there a truck?' in a street scene than it is to answer `Is there a couch?'

For \textbf{sport recognition} questions, we detect the presence of specific sports equipment in the annotations and ask questions about the type of sport being played. Images must only contain sports equipment for one particular sport. A similar approach was used to create scene understanding questions.  For example, if a toilet and a sink are present in annotations, the room is a bathroom and an appropriate scene recognition question can be created. Additionally, we use the supercategories `indoor' and `outdoor' to ask questions about where a photo was taken.

For creating \textbf{positional reasoning} questions, we use the relative locations of bounding boxes to create questions similar to  `What is to the left/right of \texttt{<object>}?' This can be ambiguous due to overlapping objects, so we employ the following heuristics to eliminate ambiguity: 1) The vertical separation between the two bounding boxes should be within a small threshold; 2) The objects should not overlap by more than the half the length of its counterpart; and 3) The objects should not be horizontally separated by more than a distance threshold, determined by subjectively judging optimal separation to reduce ambiguity. We tried to generate above/below questions, but the results were unreliable.

\textbf{Absurd questions} test the ability of an algorithm to judge when a question is not answerable based on the image's content. To make these, we make a list of the objects that are absent from a given image, and then we find questions from rest of TDIUC that ask about these absent objects, with the exception of yes/no and counting questions. This includes questions imported from COCO-VQA, auto-generated questions, and manually created questions. We make a list of all possible questions that would be `absurd' for each image and we uniformly sample three questions per image. In effect, we will have same question repeated multiple times throughout the dataset, where it can either be a genuine question or a nonsensical question. The algorithm must answer `Does Not Apply' if the question is absurd.

\subsubsection{Questions Using Visual Genome annotations}
Visual Genome's annotations contain region descriptions, relationship graphs, and object boundaries. However, the annotations can be both non-exhaustive and duplicated, which makes using them to automatically make QA pairs difficult. We only use Visual Genome to make color and positional reasoning questions. The methods we used are similar to those used with COCO, but additional precautions were needed due to quirks in their annotations. Additional details are provided in the Appendix.

\subsection{Manual Annotation}

Creating sentiment understanding and object utility/affordance questions cannot be readily done using templates, so we used manual annotation to create these. Twelve volunteer annotators were trained to generate these questions, and they used a web-based annotation tool that we developed. They were shown random images from COCO and Visual Genome and could also upload images.
\subsection{Post Processing}

Post processing was performed on questions from all sources. All numbers were converted to text, e.g., 2 became two. All answers were converted to lowercase, and trailing punctuation was stripped. Duplicate questions for the same image were removed. All questions had to have answers that appeared at least twice. The dataset was split into train and test splits with 70\% for train and 30\% for test.

\begin{table}
\centering
\footnotesize
\caption{The number of questions per type in TDIUC.}
\label{tab:num-types}
\begin{tabular}{lrc}
\toprule
                        &\textbf{Questions} & \textbf{Unique Answers}  \\ \midrule
Scene Recognition       	& 66,706          & 83               \\
Sport Recognition       	& 31,644          & 12              \\
Color Attributes        	& 195,564         & 16              \\
Other Attributes        	& 28,676          & 625            \\
Activity Recognition    	& 8,530           & 13             \\
Positional Reasoning    	& 38,326          & 1,300           \\
Sub. Object Recognition     & 93,555          & 385               \\
Absurd                  	& 366,654         & 1               \\
Utility/Affordance      	& 521             & 187            \\
Object Presence         	& 657,134         & 2               \\
Counting                	& 164,762         & 16            \\
Sentiment Understanding 	& 2,095           & 54               \\ \midrule
Grand Total             	& 1,654,167       & 1,618           \\ \bottomrule
\end{tabular}
\end{table}

\section{Proposed Evaluation Metric}
\label{sec:proposed-metric}

One of the main goals of VQA research is to build computer vision systems capable of many tasks, instead of only having expertise at one specific task (e.g., object recognition). For this reason, some have argued that VQA is a kind of Visual Turing Test~\cite{malinowski2014multi}. However, if simple accuracy is used for evaluating performance, then it is hard to know if a system succeeds at this goal because some question-types have far more questions than others. In VQA, skewed distributions of question-types are to be expected. If each test question is treated equally, then it is difficult to assess performance on rarer question-types and to compensate for bias. We propose multiple measures to compensate for bias and skewed distributions.

To compensate for the skewed question-type distribution, we compute accuracy for each of the 12 question-types separately. However, it is also important to have a final unified accuracy metric. Our overall metrics are the arithmetic and harmonic means across all per question-type accuracies, referred to as arithmetic mean-per-type (Arithmetic MPT) accuracy and harmonic mean-per-type accuracy (Harmonic MPT). Unlike the Arithmetic MPT, Harmonic MPT measures the ability of a system to have high scores across all question-types and is skewed towards lowest performing categories.

We also use normalized metrics that compensate for bias in the form of imbalance in the distribution of answers within each question-type, e.g., the most repeated answer `two' covers over 35\% of all the counting-type questions. To do this, we compute the accuracy for each unique answer separately within a question-type and then average them together for the question-type. To compute overall performance, we compute the arithmetic normalized mean per-type (N-MPT) and harmonic N-MPT scores. A large discrepancy between unnormalized and normalized scores suggests an algorithm is not generalizing to rarer answers. 

\section{Algorithms for VQA}

While there are alternative formulations (e.g.,~\cite{gao2015you,jabri2016revisiting}), the majority of VQA systems formulate it as a classification problem in which the system is given an image and a question, with the answers as categories.~\cite{antol2015vqa,ren2015image,FukuiPYRDR16,shih2016,ilievski2016focused,kim2016multimodal,jiang2015compositional,LuYBP16,NohSH15,saito2016dualnet,WuWSHD15,xu2015ask,Yang2016,zhou2015,jabri2016revisiting,malinowskiRF15}. Almost all systems use CNN features to represent the image and either a recurrent neural network (RNN) or a bag-of-words model for the question. We briefly review some of these systems, focusing on the models we compare in experiments. For a more comprehensive review, see \cite{kafle2016review} and \cite{wu2016visual}.

Two simple VQA baselines are linear or multi-layer perceptron (MLP) classifiers that take as input the question and image embeddings concatenated to each other~\cite{antol2015vqa,Kafle2016,zhou2015}, where the image features come from the last hidden layer of a CNN. These simple approaches often work well and can be competitive with complex attentive models~\cite{Kafle2016,zhou2015}.

Spatial attention has been heavily investigated in VQA models~\cite{FukuiPYRDR16, shih2016, Yang2016, xiong2016dynamic,xu2015ask,LuYBP16,ilievski2016focused}. These systems weigh the visual features based on their relevance to the question, instead of using global features, e.g., from the last hidden layer of a CNN. For example, to answer `What color is the bear?' they aim emphasize the visual features around the bear and suppress other features. 

The MCB system~\cite{FukuiPYRDR16} won the CVPR-2016 VQA Workshop Challenge. In addition to using spatial attention, it implicitly computes the outer product between the image and question features to ensure that all of their elements interact. Explicitly computing the outer product would be slow and extremely high dimensional, so it is done using an efficient approximation. It uses an long short-term memory (LSTM) networks to embed the question. 

The neural module network (NMN) is an especially interesting compositional approach to VQA~\cite{AndreasRDK15,andreas2016learning}. The main idea is to compose a series of discrete modules (sub-networks) that can be executed collectively to answer a given question. To achieve this, they use a variety of modules, e.g., the \texttt{find(x)} module outputs a heat map for detecting $x$. To arrange the modules, the question is first parsed into a concise expression (called an S-expression), e.g., `What is to the right of the car?' is parsed into \texttt{(what car);(what right);(what (and car right))}. Using these expressions, modules are composed into a sequence to answer the query.

The multi-step recurrent answering units (RAU) model for VQA is another state-of-the-art method~\cite{noh2016training}. Each inference step in RAU consists of a complete answering block that takes in an image, a question, and the output from the previous LSTM step. Each of these is part of a larger LSTM network that progressively reasons about the question. 

\section{Experiments}\label{sec:experiments}
\begin{table*}
\centering
\footnotesize
\caption{Results for all VQA models. The unnormalized accuracy for each question-type is shown. Overall performance is reported using 5 metrics. Overall (Arithmetic MPT) and Overall (Harmonic MPT) are averages of these sub-scores, providing a clearer picture of performance across question-types than simple accuracy. Overall Arithmetic N-MPT and Harmonic N-MPT normalize across unique answers to better analyze the impact of answer imbalance (see Sec.~\ref{sec:proposed-metric}). Normalized scores for individual question-types are presented in the appendix table \ref{tbl:results-appendix}. * denotes training without absurd questions.}
\label{tbl:results}
\begin{tabular}{@{}lrrrrrrrrrrrr@{}}
\toprule
                        &\textbf{YES} & \textbf{REP} & \textbf{IMG} & \textbf{QUES} & \textbf{Q+I} & \textbf{*Q+I}  & \textbf{MLP}   & \textbf{MCB} & \textbf{*MCB} & \textbf{MCB-A} & \textbf{NMN} & \textbf{RAU}\\ \midrule
Scene Recognition        		& 26.90 & 26.90				& 14.25 & 53.18   & 72.19 & 72.75 & 91.45 & 92.04 & 91.87 & 93.06    &  91.88   &  \textbf{93.96}   \\
Sport Recognition        		& 0.00  & 22.05   			& 18.61 & 18.87  & 85.16 & 89.40 & 90.24 & 92.47 & 92.47 & 92.77   &  89.99   &  \textbf{93.47}  \\
Color Attributes         		& 0.00  & 22.74   			& 0.92 & 37.60   & 43.69 & 50.52 & 53.64 & 56.93 & 57.07 & \textbf{68.54}   &  54.91   &  66.86  \\
Other Attributes          		& 0.00  & 24.23   			& 2.07 & 36.13   & 42.89 & 51.47 & 41.79 & 53.24 & 54.62 & \textbf{56.72}   &  47.66   &  56.49  \\
Activity Recognition     		& 0.00  & 21.63   			& 3.06 & 10.81    & 24.16 & 48.55 & 39.22 & 51.42 & \textbf{53.58} & 52.35   &  44.26   &  51.60  \\
Positional Reasoning     		& 0.00  & 6.05    			& 2.23 & 14.23   & 25.15 & 27.73 & 21.87 & 33.34 & 33.02 & \textbf{35.40}   &  27.92   &  35.26  \\
Sub. Object Recognition 		& 0.00  & 7.16    			& 10.55 & 21.40   & 80.92 & 81.66 & 80.55 & 84.63 & 84.58 & 85.54   &  82.02   &  \textbf{86.11}  \\
Absurd                   		& 0.00  & \textbf{100.00}  	& 19.97 & 96.71    & 96.98 & N/A   & 95.96 & 83.44 & N/A   & 84.82   &  87.51   &  96.08  \\
Utility and Affordances   		& 11.70  & 11.70   			& 5.26 & 16.37      & 24.56 & 30.99 & 13.45 & 33.92 & 29.24 & \textbf{35.09}   &  25.15   &  31.58  \\
Object Presence          		& 50.00 & 50.00   			& 20.73 & 69.06   & 69.43 & 69.50 & 92.33 & 91.84 & 91.55 & 93.64   &  92.50   &  \textbf{94.38}  \\
Counting                 		& 0.00  & 36.19   			& 0.30 & 44.51   & 44.82 & 44.84 & 51.12 & 50.29 & 50.07 & \textbf{51.01}   &  49.21   &  48.43  \\
Sentiment Understanding  		& 44.64  & 44.64   			& 15.93 & 52.84     & 53.00 & 59.94 & 58.33 & 65.46 & 66.25 & \textbf{66.25}   &  58.04   &  60.09  \\ \midrule
Overall (Arithmetic MPT)		& 11.10  & 31.11   			& 9.49 & 39.31      & 55.25 & 57.03 & 60.87 & 65.75 & 66.07 & \textbf{67.90} &  62.59   &  67.81  \\
Overall (Harmonic MPT)  		& 0.00  & 17.53   			& 1.92 & 25.93     & 44.13 & 50.30 & 42.80 & 58.03 & 55.43 & \textbf{60.47}   &  51.87   &  59.00  \\ \midrule
Overall (Arithmetic N-MPT)  	& 4.87  & 15.63   			& 5.82 & 21.46      & 29.47 & 28.10 & 31.36 & 39.81 & 35.49 & \textbf{42.24}  &  34.00   &  41.04  \\
Overall (Harmonic N-MPT)  		& 0.00  & 0.83    			& 1.91 & 8.42      & 14.99 & 18.30 & 9.46  & 24.77 & 23.20 & \textbf{27.28}   &  16.67   &  23.99   \\\midrule
Simple Accuracy          		& 21.14 & 51.15   			& 14.54 & 62.74   & 69.53 & 63.30 & 81.07 & 79.20 & 78.06 & 81.86   &  79.56   &  \textbf{84.26}    \\ \bottomrule
\end{tabular}
\end{table*}

We trained multiple baseline models as well as state-of-the-art VQA methods on TDIUC. The methods we use are:
\begin{itemize}[noitemsep,nolistsep]
\item \textbf{YES}: Predicts `yes' for all questions.
\item \textbf{REP}: Predicts the most repeated answer in a question-type category using an oracle.
\item \textbf{QUES}: A linear softmax classifier given only question features (image blind). 
\item \textbf{IMG}: A linear softmax classifier  given only image features (question blind).
\item \textbf{Q+I}: A linear classifier given the question and image..
\item \textbf{MLP}: A 4-layer MLP fed question and image features.
\item \textbf{MCB}: MCB~\cite{FukuiPYRDR16} without spatial attention.
\item \textbf{MCB-A}: MCB~\cite{FukuiPYRDR16} with spatial attention.
\item \textbf{NMN}: NMN from \cite{AndreasRDK15} with minor modifications.
\item \textbf{RAU}: RAU~\cite{noh2016training} with minor modifications.

\end{itemize}
For image features, ResNet-152~\cite{resnet} with $448 \times 448$ images was used for all models.

QUES and IMG provide information about biases in the dataset. QUES, Q+I, and MLP all use 4800-dimensional skip-thought vectors~\cite{kiros2015skip} to embed the question, as was done in \cite{Kafle2016}. For image features, these all use the `pool5' layer of ResNet-152 normalized to unit length. MLP is a 4-layer net with a softmax output layer. The 3 ReLU hidden layers have 6000, 4000, and 2000 units, respectively. During training, dropout (0.3) was used for the hidden layers.

For MCB, MCB-A, NMN and RAU, we used publicly available code to train them on TDIUC. The experimental setup and hyperparamters were kept unchanged from the default choices in the code, except for upgrading NMN and RAU's visual representation to both use ResNet-152.

Results on TDIUC for these models are given in Table~\ref{tbl:results}. Accuracy scores are given for each of the 12 question-types in Table~\ref{tbl:results}, and scores that are normalized by using mean-per-unique-answer are given in appendix Table \ref{tbl:results-appendix}.
\section{Detailed Analysis of VQA Models}
\subsection{Easy Question-Types for Today's Methods}
By inspecting Table~\ref{tbl:results}, we can see that some question-types are comparatively easy ($>90$\%) under MPT: scene recognition, sport recognition, and object presence. High accuracy is also achieved on absurd, which we discuss in greater detail in Sec.~\ref{sec:absurd}. Subordinate object recognition is moderately high ($>80$\%), despite having a large number of unique answers. Accuracy on counting is low across all methods, despite a large number of training data. For the remaining question-types, more analysis is needed to pinpoint whether the weaker performance is due to lower amounts of training data, bias, or limitations of the models. We next investigate how much of the good performance is due to bias in the answer distribution, which N-MPT compensates for.

\subsection{Effects of the Proposed Accuracy Metrics}

One of our major aims was to compensate for the fact that algorithms can achieve high scores by simply learning to answer more populated and easier question-types. For existing datasets, earlier work has shown that simple baseline methods routinely exceed more complex methods using simple accuracy~\cite{Kafle2016,zhou2015,jabri2016revisiting}. On TDIUC, MLP surpasses MCB and NMN in terms of simple accuracy, but a closer inspection reveals that MLP's score is highly determined by performance on categories with a large number of examples, such as `absurd' and `object presence.' Using MPT, we find that both NMN and MCB outperform MLP. Inspecting normalized scores for each question-type (Appendix Table \ref{tbl:results-appendix}) shows an even more pronounced differences, which is also reflected in arithmetic N-MPT score presented in Table~\ref{tbl:results}. This indicates that MLP is prone to overfitting. Similar observations can be made for MCB-A compared to RAU, where RAU outperforms MCB-A using simple accuracy, but scores lower on \textit{all} the metrics designed to compensate for the skewed answer distribution and bias.

Comparing the unnormalized and normalized metrics can help us determine the generalization capacity of the VQA algorithms for a given question-type. A large difference in these scores suggests that an algorithm is relying on the skewed answer distribution to obtain high scores. We found that for MCB-A, the accuracy on subordinate object recognition drops from 85.54\% with unnormalized to 23.22\% with normalized, and for scene recognition it drops from 93.06\% (unnormalized) to 38.53\% (normalized). Both these categories have a heavily skewed answer distribution; the top-25 answers in subordinate object recognition and the top-5 answers in scene recognition cover over 80\% of all questions in their respective question-types. This shows that question-types that appear to be easy may simply be due to the algorithms learning the answer statistics. A truly easy question-type will have similar performance for both unnormalized and normalized metrics. For example, sport recognition shows only 17.39\% drop compared to a 30.21\% drop for counting, despite counting having same number of unique answers and far more training data. By comparing relative drop in performance between normalized and unnormalized metric, we can also \textit{compare} the generalization capability of the algorithms, e.g., for subordinate object recognition, RAU has higher unnormalized score (86.11\%) compared to MCB-A (85.54\%). However, for normalized scores, MCB-A  has significantly higher performance (23.22\%) than RAU (21.67\%). This shows RAU may be more dependent on the answer distribution.  Similar observations can be made for MLP compared to MCB.

\subsection{Can Algorithms Predict Rare Answers?}
\label{sec:rare}
In the previous section, we saw that the VQA models struggle to correctly predict rarer answers. Are the less repeated questions \textit{actually} harder to answer, or are the algorithms simply biased toward more frequent answers? To study this, we created a subset of TDIUC that only consisted of questions that have answers repeated less than 1000 times. We call this dataset TDIUC-Tail, which has 46,590 train and 22,065 test questions. Then, we trained MCB on: 1) the full TDIUC dataset; and 2) TDIUC-Tail. Both versions were evaluated on the validation split of TDIUC-Tail.

We found that MCB trained only on TDIUC-Tail outperformed MCB trained on all of TDIUC across all question-types (details are in appendix Table \ref{tbl:results-umpt-ablation} and \ref{tbl:results-nmpt-ablation}). This shows that MCB is capable of learning to correctly predict rarer answers, but it is simply biased towards predicting more common answers to maximize overall accuracy. Using normalized accuracy disincentivizes the VQA algorithms' reliance on the answer statistics, and for deploying a VQA system it may be useful to optimize directly for N-MPT.

\subsection{Effects of Including Absurd Questions}
\label{sec:absurd}
Absurd questions force a VQA system to look at the image to answer the question. In TDIUC, these questions are sampled from the rest of the dataset, and they have a high prior probability of being answered `Does not apply.' This is corroborated by the QUES model, which achieves a high accuracy on absurd; however, for the same questions when they are genuine for an image, it only achieves 6.77\% accuracy on these questions. Good absurd performance is achieved by sacrificing performance on other categories. A robust VQA system should be able to detect absurd questions without then failing on others. By examining the accuracy on real questions that are identical to absurd questions, we can quantify an algorithm's ability to differentiate the absurd questions from the real ones. We found that simpler models had much lower accuracy on these questions, (QUES: 6.77\%, Q+I: 34\%), compared to more complex models (MCB: 62.44\%, MCB-A: 68.83\%).

To further study this, we we trained two VQA systems, Q+I and MCB, both with and without absurd. The results are presented in Table~\ref{tbl:results}. For Q+I trained without absurd questions, accuracies for other categories increase considerably compared to Q+I trained with full TDIUC, especially for question-types that are used to sample absurd questions, e.g., activity recognition (24\% when trained with absurd and 48\% without). Arithmetic MPT accuracy for the Q+I model that is trained without absurd (57.03\%) is also substantially greater than MPT for the model trained with absurd (51.45\% for all categories except absurd). This suggests that Q+I is not properly discriminating between absurd and real questions and is biased towards mis-identifying genuine questions as being absurd. In contrast, MCB, a more capable model, produces worse results for absurd, but the version trained without absurd shows much smaller differences than Q+I, which shows that MCB is more capable of identifying absurd questions.

\subsection{Effects of Balancing Object Presence}

In Sec.~\ref{sec:rare}, we saw that a skewed answer distribution can impact generalization. This effect is strong even for simple questions and affects even the most sophisticated algorithms. Consider MCB-A when it is trained on both COCO-VQA and Visual Genome, i.e., the winner of the CVPR-2016 VQA Workshop Challenge. When it is evaluated on object presence questions from TDIUC, which contains 50\% `yes' and 50\% `no' questions, it correctly predicts `yes' answers with 86.3\% accuracy, but only 11.2\% for questions with `no' as an answer. However, after training it on TDIUC, MCB-A is able to achieve 95.02\% for `yes' and 92.26\% for `no.' MCB-A performed poorly by learning the biases in the COCO-VQA dataset, but it is capable of performing well when the dataset is unbiased. Similar observations about balancing yes/no questions were made in \cite{Zhang_2016_CVPR}. Datasets could balance simple categories like object presence, but extending the same idea to all other categories is a challenging task and undermines the natural statistics of the real-world. Adopting mean-per-class and normalized accuracy metrics can help compensate for this problem.

\subsection{Advantages of Attentive Models}

By breaking questions into types, we can assess which types benefit the most from attention. We do this by comparing the MCB model with and without attention, i.e., MCB and MCB-A. As seen in Table~\ref{tbl:results}, attention helped improve results on several question categories. The most pronounced increases are for color recognition, attribute recognition, absurd, and counting. All of these question-types require the algorithm to detect specified object(s) (or lack thereof) to be answered correctly. MCB-A computes attention using local features from different spatial locations, instead of global image features. This aids in localizing individual objects. The attention mechanism learns the relative importance of these features. RAU also utilizes spatial attention and shows similar increments.

\subsection{Compositional and Modular Approaches}

NMN, and, to a lesser extent, RAU propose compositional approaches for VQA. For COCO-VQA, NMN has performed worse than some MLP models~\cite{Kafle2016} using simple accuracy. We hoped that it would achieve better performance than other models for questions that require logically analyzing an image in a step-by-step manner, e.g., positional reasoning. However, while NMN did perform better than MLP using MPT and N-MPT metric, we did not see any substantial benefits in specific question-types. This may be because NMN is limited by the quality of the `S-expression' parser, which produces incorrect or misleading parses in many cases.  For example, `What color is the jacket of the man on the far left?' is parsed as \texttt{(color jacket);(color leave);(color (and jacket leave))}. This expression not only fails to parse `the man', which is a crucial element needed to correctly answer the question, but also wrongly interprets `left' as past tense of leave. 

RAU performs inference over multiple hops, and because each hop contains a complete VQA system, it can learn to solve different tasks in each step. Since it is trained end-to-end, it does not need to rely on rigid question parses. It showed very good performance in detecting absurd questions and also performed well on other categories.

\section{Conclusion}

We introduced TDIUC, a VQA dataset that consists of 12 explicitly defined question-types, including absurd questions, and we used it to perform a rigorous analysis of recent VQA algorithms. We proposed new evaluation metrics to compensate for biases in VQA datasets. Results show that the absurd questions and the new evaluation metrics enable a deeper understanding of VQA algorithm behavior. 

{\small
\bibliographystyle{ieee}
\bibliography{egbib}

\begin{thebibliography}{10}\itemsep=-1pt

\bibitem{AndreasRDK15}
J.~Andreas, M.~Rohrbach, T.~Darrell, and D.~Klein.
\newblock Deep compositional question answering with neural module networks.
\newblock In {\em CVPR}, 2016.

\bibitem{andreas2016learning}
J.~Andreas, M.~Rohrbach, T.~Darrell, and D.~Klein.
\newblock Learning to compose neural networks for question answering.
\newblock In {\em NAACL}, 2016.

\bibitem{antol2015vqa}
S.~Antol, A.~Agrawal, J.~Lu, M.~Mitchell, D.~Batra, C.~L. Zitnick, and
  D.~Parikh.
\newblock {VQA}: {V}isual question answering.
\newblock In {\em ICCV}, 2015.

\bibitem{Fei-Fei2006}
L.~Fei-Fei, R.~Fergus, and P.~Perona.
\newblock One-shot learning of object categories.
\newblock {\em IEEE Trans. Pattern Analysis and Machine Intelligence},
  28:594--611, 2006.

\bibitem{FukuiPYRDR16}
A.~Fukui, D.~H. Park, D.~Yang, A.~Rohrbach, T.~Darrell, and M.~Rohrbach.
\newblock Multimodal compact bilinear pooling for visual question answering and
  visual grounding.
\newblock In {\em EMNLP}, 2016.

\bibitem{gao2015you}
H.~Gao, J.~Mao, J.~Zhou, Z.~Huang, L.~Wang, and W.~Xu.
\newblock Are you talking to a machine? {D}ataset and methods for multilingual
  image question answering.
\newblock In {\em NIPS}, 2015.

\bibitem{balanced_vqa_v2}
Y.~Goyal, T.~Khot, D.~Summers{-}Stay, D.~Batra, and D.~Parikh.
\newblock Making the {V} in {VQA} matter: Elevating the role of image
  understanding in {V}isual {Q}uestion {A}nswering.
\newblock In {\em CVPR}, 2017.

\bibitem{resnet}
K.~He, X.~Zhang, S.~Ren, and J.~Sun.
\newblock Deep residual learning for image recognition.
\newblock In {\em CVPR}, 2016.

\bibitem{ilievski2016focused}
I.~Ilievski, S.~Yan, and J.~Feng.
\newblock A focused dynamic attention model for visual question answering.
\newblock {\em arXiv preprint arXiv:1604.01485}, 2016.

\bibitem{jabri2016revisiting}
A.~Jabri, A.~Joulin, and L.~van~der Maaten.
\newblock Revisiting visual question answering baselines.
\newblock In {\em ECCV}, 2016.

\bibitem{jiang2015compositional}
A.~Jiang, F.~Wang, F.~Porikli, and Y.~Li.
\newblock Compositional memory for visual question answering.
\newblock {\em arXiv preprint arXiv:1511.05676}, 2015.

\bibitem{johnson2016clevr}
J.~Johnson, B.~Hariharan, L.~van~der Maaten, L.~Fei-Fei, C.~L. Zitnick, and
  R.~Girshick.
\newblock {CLEVR}: A diagnostic dataset for compositional language and
  elementary visual reasoning.
\newblock In {\em CVPR}, 2017.

\bibitem{Kafle2016}
K.~Kafle and C.~Kanan.
\newblock Answer-type prediction for visual question answering.
\newblock In {\em CVPR}, 2016.

\bibitem{kafle2016review}
K.~Kafle and C.~Kanan.
\newblock Visual question answering: Datasets, algorithms, and future
  challenges.
\newblock {\em Computer Vision and Image Understanding}, 2017.

\bibitem{kafle2017b}
K.~Kafle, M.~Yousefhussien, and C.~Kanan.
\newblock Data augmentation for visual question answering.
\newblock In {\em International Conference on Natural Language Generation
  (INLG)}, 2017.

\bibitem{kim2016multimodal}
J.-H. Kim, S.-W. Lee, D.-H. Kwak, M.-O. Heo, J.~Kim, J.-W. Ha, and B.-T. Zhang.
\newblock Multimodal residual learning for visual {QA}.
\newblock In {\em NIPS}, 2016.

\bibitem{kiros2015skip}
R.~Kiros, Y.~Zhu, R.~Salakhutdinov, R.~S. Zemel, A.~Torralba, R.~Urtasun, and
  S.~Fidler.
\newblock Skip-thought vectors.
\newblock In {\em NIPS}, 2015.

\bibitem{krishnavisualgenome}
R.~Krishna, Y.~Zhu, O.~Groth, J.~Johnson, K.~Hata, J.~Kravitz, S.~Chen,
  Y.~Kalantidis, L.-J. Li, D.~A. Shamma, M.~Bernstein, and L.~Fei-Fei.
\newblock {Visual Genome}: Connecting language and vision using crowdsourced
  dense image annotations.
\newblock 2016.

\bibitem{lin2014microsoft}
T.-Y. Lin, M.~Maire, S.~Belongie, J.~Hays, P.~Perona, D.~Ramanan,
  P.~Doll{\'a}r, and C.~L. Zitnick.
\newblock Microsoft {COCO}: Common objects in context.
\newblock In {\em ECCV}. 2014.

\bibitem{LuYBP16}
J.~Lu, J.~Yang, D.~Batra, and D.~Parikh.
\newblock Hierarchical question-image co-attention for visual question
  answering.
\newblock In {\em NIPS}, 2016.

\bibitem{malinowski2014multi}
M.~Malinowski and M.~Fritz.
\newblock A multi-world approach to question answering about real-world scenes
  based on uncertain input.
\newblock In {\em NIPS}, 2014.

\bibitem{malinowskiRF15}
M.~Malinowski, M.~Rohrbach, and M.~Fritz.
\newblock Ask your neurons: {A} neural-based approach to answering questions
  about images.
\newblock In {\em ICCV}, 2015.

\bibitem{noh2016training}
H.~Noh and B.~Han.
\newblock Training recurrent answering units with joint loss minimization for
  {VQA}.
\newblock {\em arXiv preprint arXiv:1606.03647}, 2016.

\bibitem{NohSH15}
H.~Noh, P.~H. Seo, and B.~Han.
\newblock Image question answering using convolutional neural network with
  dynamic parameter prediction.
\newblock In {\em CVPR}, 2016.

\bibitem{ren2015image}
M.~Ren, R.~Kiros, and R.~Zemel.
\newblock Exploring models and data for image question answering.
\newblock In {\em NIPS}, 2015.

\bibitem{saito2016dualnet}
K.~Saito, A.~Shin, Y.~Ushiku, and T.~Harada.
\newblock Dualnet: Domain-invariant network for visual question answering.
\newblock {\em arXiv preprint arXiv:1606.06108}, 2016.

\bibitem{shih2016}
K.~J. Shih, S.~Singh, and D.~Hoiem.
\newblock Where to look: Focus regions for visual question answering.
\newblock In {\em CVPR}, 2016.

\bibitem{wu2016visual}
Q.~Wu, D.~Teney, P.~Wang, C.~Shen, A.~Dick, and A.~v.~d. Hengel.
\newblock Visual question answering: A survey of methods and datasets.
\newblock {\em Computer Vision and Image Understanding}, 2017.

\bibitem{WuWSHD15}
Q.~Wu, P.~Wang, C.~Shen, A.~van~den Hengel, and A.~R. Dick.
\newblock Ask me anything: Free-form visual question answering based on
  knowledge from external sources.
\newblock In {\em CVPR}, 2016.

\bibitem{xiong2016dynamic}
C.~Xiong, S.~Merity, and R.~Socher.
\newblock Dynamic memory networks for visual and textual question answering.
\newblock In {\em ICML}, 2016.

\bibitem{xu2015ask}
H.~Xu and K.~Saenko.
\newblock Ask, attend and answer: Exploring question-guided spatial attention
  for visual question answering.
\newblock In {\em ECCV}, 2016.

\bibitem{Yang2016}
Z.~Yang, X.~He, J.~Gao, L.~Deng, and A.~J. Smola.
\newblock Stacked attention networks for image question answering.
\newblock In {\em CVPR}, 2016.

\bibitem{Zhang_2016_CVPR}
P.~Zhang, Y.~Goyal, D.~Summers-Stay, D.~Batra, and D.~Parikh.
\newblock Yin and yang: Balancing and answering binary visual questions.
\newblock In {\em CVPR}, 2016.

\bibitem{zhou2015}
B.~Zhou, Y.~Tian, S.~Sukhbaatar, A.~Szlam, and R.~Fergus.
\newblock Simple baseline for visual question answering.
\newblock {\em CoRR}, abs/1512.02167, 2015.

\bibitem{visual7w}
Y.~Zhu, O.~Groth, M.~Bernstein, and L.~Fei-Fei.
\newblock Visual7w: Grounded question answering in images.
\newblock In {\em CVPR}, 2016.

\end{thebibliography}
}

\begin{appendices}

\section{Additional Details About TDIUC}

In this section, we will provide additional details about the TDIUC dataset creation and additional statistics that were omitted from the main paper due to inadequate space.

\subsection{Questions using Visual Genome Annotations}

As mentioned in the main text, Visual Genome's annotations are both non-exhaustive and duplicated. This makes using them to automatically make question-answer (QA) pairs difficult. Due to these issues, we only used them to make two types of questions: Color Attributes and Positional Reasoning. Moreover, a number of restrictions needed to be placed, which are outlined below.

For making Color Attribute questions, we make use of the attributes metadata in the Visual Genome annotations to populate the template `What color is the \texttt{<object>}?' However, Visual Genome metadata can contain several color attributes for the same object as well as different names for the same object. Since the annotators type the name of the object manually rather than choosing from a predetermined set of objects, the same object can be referred by different names, e.g., `xbox controller,' `game controller,' `joystick,' and `controller' can all refer to same object in an image. The object name is sometimes also accompanied by its color, e.g., `white horse' instead of `horse' which makes asking the Color Attribute question `What color is the white horse?' pointless. One potential solution is to use the wordnet `synset' which accompanies every object annotation in the Visual Genome annotations. Synsets are used to group different variations of the common objects names under a single noun from wordnet. However, we found that the synset matching was erroneous in numerous instances, where the object category was misrepresented by the given synset. For example, A `controller' is matched with synset `accountant' even when the `controller' is referring to a game controller. Similarly, a `cd' is matched with synset of `cadmium.' To avoid these problems we made a set of stringent requirements before making questions:
\begin{enumerate}
\item The chosen object should only have a single attribute that belongs to a set of commonly used colors.
\item The chosen object name or synset must be one of the 91 common objects in the MS-COCO annotations.
\item There must be only one instance of the chosen object.
\end{enumerate}
Using these criteria, we found that we could safely ask the question of the form `What color is the \texttt{<object>}?'.
\begin{table}
\centering
\caption{The number of questions produced via each source.}
\footnotesize
\label{tab:question-sources}
\begin{tabular}{lrrc}
\toprule
  & \textbf{Questions} & \textbf{Images} & \textbf{Unique Answers} \\ \midrule

Imported (VQA)     & 49,990      & 43,636   & 812   \\
Imported (Genome)  & 310,225     & 89,039   & 1,446   \\
Generated (COCO)   & 1,286,624   & 122,218  & 108    \\
Generated (Genome) & 6,391       & 5,988    & 675      \\
Manual          & 937         & 740      & 218       \\ \midrule
Grand Total     & 1,654,167        &  167,437      & 1,618                            \\ \bottomrule
\end{tabular}
\end{table}

Similarly, for making Positional Reasoning questions, we used the relationships metadata in the Visual Genome annotations. The relationships metadata connects two objects by a relationship phrase. Many of these relationships describe the positions of the two objects, e.g., A is `on right' of B, where `on right' is one of the example relationship clause from Visual Genome, with the object A as the subject and the object B as the object. This can be used to generate Positional Reasoning questions. Again, we take several measures to avoid ambiguity. First, we only use objects that appear once in the image because `What is to the left of A' can be ambiguous if there are two instances of the object A. However, since visual genome annotations are non-exhaustive, there may still (rarely) be more than one instance of object A that was not annotated. To disambiguate such cases, we use the attributes metadata to further specify the object wherever possible, e.g., instead of asking `What is to the right of the bus?', we ask `What is to the right of the green bus?'

Due to a these stringent criteria, we could only create a small number of questions using Visual Genome annotations compared to other sources. The number of questions produced via each source is shown in Table \ref{tab:question-sources}.

\begin{figure*}
\centering
\includegraphics[width=\textwidth]{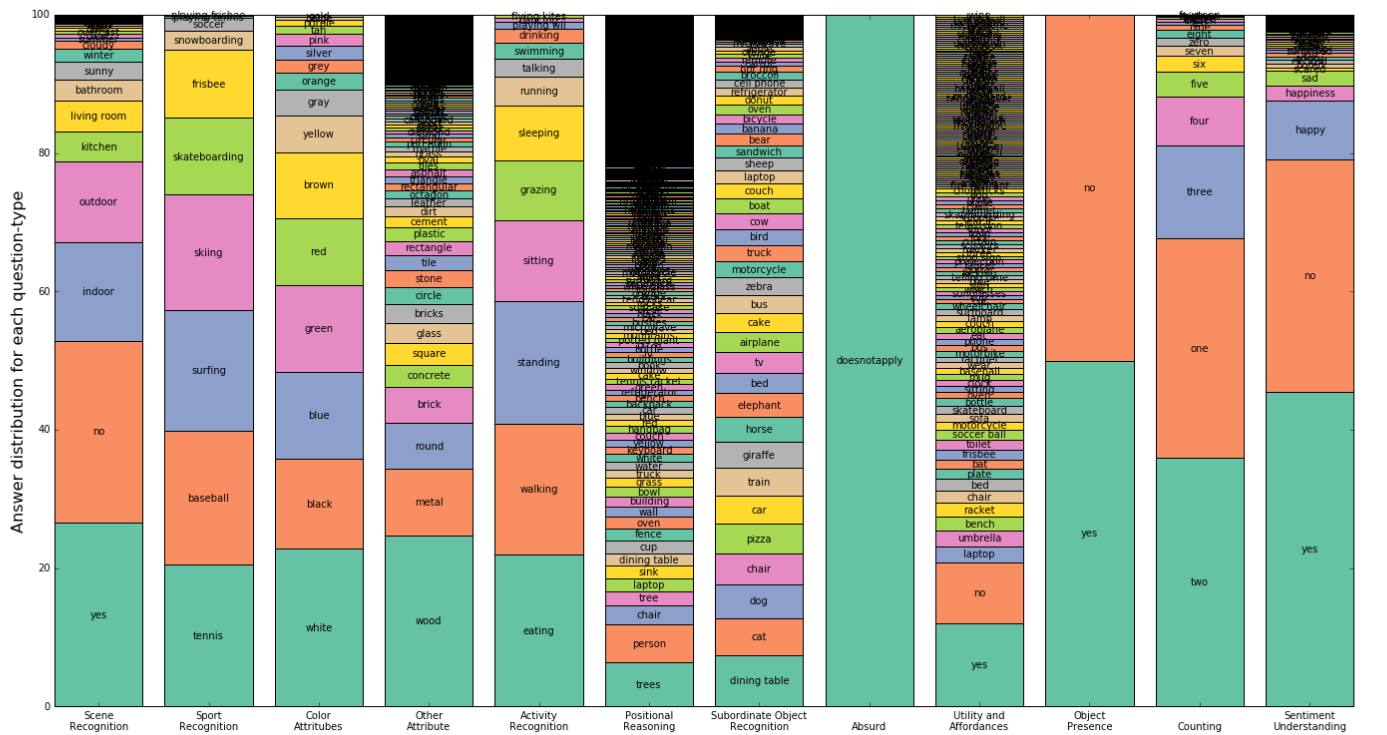}\\
\caption{Answer distributions for the answers for each of the question-types. This shows the relative frequency of each unique answer within a question-type, so for some question-types, e.g., counting, even slim bars contain a fairly large number of instances with that answer. Similarly, for less populated question-types such as utility and affordances, even large bars represents only a small number of training examples.}
\label{tbl:distribution} 
\end{figure*}

\subsection{Answer Distribution}
Figure \ref{tbl:distribution} shows the answer distribution for the different question-types. We can see that some categories, such as counting, scene recognition and sentiment understanding, have a very large share of questions represented by only a few top answers. In such cases, the performance of a VQA algorithm can be inflated unless the evaluation metric compensates for this bias. In other cases, such as positional reasoning and object utility and affordances, the answers are much more varied, with top-50 answers covering less than 60\% of all answers. 

We have completely balanced answer distribution for object presence questions, where exactly 50\% of questions being answered `yes' and the remaining 50\% of the questions are answered `no'. For other categories, we have tried to design our question generation algorithms so that a single answer does not have a significant majority within a question type. For example, while scene understanding has top-4 answers covering over 85\% of all the questions, there are roughly as many `no' questions (most common answer) as there are `yes' questions (second most-common answer). Similar distributions can be seen for counting, where `two' (most-common answer) is repeated almost as many times as `one' (second most-common answer). By having at least the top-2 answers split almost equally, we remove the incentive for an algorithm to perform well using simple mode guessing, even when using the simple accuracy metric. 

\subsection{Train and Test Split}
In the paper, we mentioned that we split the entire collection into 70\% train and 30\% test/validation. To do this, we not only need to have a roughly equal distribution of question types and answers, but also need to make sure that the multiple questions for same image do not end up in two different splits, i.e., the same image cannot occur in both the train and the test partitions.  So, we took following measures to split the questions into train-test splits. First, we split all the images into three separate clusters.
\begin{enumerate}[noitemsep]
\item Manually uploaded images, which includes all the images manually uploaded by our volunteer annotators. 
\item Images from the COCO dataset, including all the images for questions generated from COCO annotations and those imported from COCO-VQA dataset. In addition, a large number of Visual Genome questions also refer to COCO images. So, some questions that are generated and imported from Visual Genome are also included in this cluster.
\item Images exclusively in the Visual Genome dataset, which includes images for a part of the questions imported from Visual Genome and those generated using that dataset.
\end{enumerate}
We follow simple rules to split each of these clusters of images into either belonging to the train or test splits. 
\begin{enumerate}[noitemsep]
\item All the questions belonging to images coming from the `train2014' split of COCO images are assigned to the train split and all the questions belonging to images from the `val2014' split are assigned to test split.
\item For manual and Visual Genome images, we randomly split 70\% of images to train and rest to test. 
\end{enumerate}

\begin{table*}
\centering
\footnotesize
\caption{Results for all the VQA models. The normalized accuracy for each question-type is shown here. The models are identical to the ones in Table 3 in main paper. Overall performance is, again, reported using all 5 metrics. Overall (Arithmetic N-MPT) and Overall (Harmonic N-MPT) are averages of the reported sub-scores. Similarly, Arithmetic MPT and Harmonic MPT are averages of sub-scores reported in Table 3 in the main paper. * denotes training without absurd questions.}
\label{tbl:results-appendix}
\begin{tabular}{@{}lrrrrrrrrrrrr@{}}
\toprule
                        &\textbf{YES} & \textbf{REP} & \textbf{IMG} & \textbf{QUES} & \textbf{Q+I} & \textbf{*Q+I}  & \textbf{MLP}   & \textbf{MCB} & \textbf{*MCB} & \textbf{MCB-A} & \textbf{NMN} & \textbf{RAU}\\ \midrule
Scene Recognition        		& 2.08 		& 2.08				& 2.83  	& 13.67   	& 25.35 		& 24.96 	& 20.54 		& 36.34 	& 32.55 			& \textbf{38.53}	&  29.06  	 	&  32.69   \\
Sport Recognition        		& 0.00  	& 9.09   			& 12.57 	& 11.09  	& 51.48 		& 60.31 	& 60.81 		& 75.25 	& 73.64 			& \textbf{75.38}	&  63.51  		&  73.60  \\
Color Attributes         		& 0.00  	& 6.25   			& 1.77 		& 20.10   	& 25.45 		& 30.37 	& 30.97 		& 36.98 	& 37.54 			& \textbf{49.40}    &  33.06   		&  46.79  \\
Other Attributes          		& 0.00  	& 0.31   			& 1.16 		& 6.21   	& 6.98  		& 9.51  	& 2.84  		& 13.90 	& 15.04 			& \textbf{15.09}    &  7.10   		&  12.11  \\
Activity Recognition     		& 0.00  	& 7.69   			& 2.88 		& 7.59    	& 16.09 		& 39.35 	& 24.95 		& 46.57 	& 48.27			 	& \textbf{48.47}	&  22.79  		&  46.65  \\
Positional Reasoning     		& 0.00  	& 0.15    			& 0.70 		& 4.03   	& 6.26  		& 8.59  	& 2.99  		& 9.29	 	& 9.39	 			& \textbf{10.76}    &  6.37   		&  9.60  \\
Sub. Object Recognition 		& 0.00  	& 0.47    			& 3.16  	& 3.72   	& 15.91 		& 16.97 	& 14.85 		& 22.07 	& 23.05 			& \textbf{23.22}	&  16.83   		&  21.67  \\
Absurd                   		& 0.00  	& \textbf{100.00}  	& 19.97	 	& 96.71    	& 96.98 		& N/A   	& 95.96 		& 83.44 	& N/A   			& 84.82 			&  87.51    	&  96.08  \\
Utility and Affordances   		& 1.22  	& 1.22   			& 1.34 		& 9.23     & 16.85 			& 21.97  	& 6.18  		& 24.07 	& 23.33 			& \textbf{26.20}    &  19.55   		&  21.38  \\
Object Presence          		& 50.00 	& 50.00   			& 20.73 	& 69.06   	& 69.43 		& 69.50 	& 92.33 		& 91.84 	& 91.95 			& 93.64   			&  92.50  		&  \textbf{94.38}  \\
Counting                 		& 0.00  	& 6.25   			& 1.31 		& 10.30   	& 14.61 		& 14.62 	& 16.43 		& 17.83 	& 18.09 			& 20.80    			&  15.52   		&  \textbf{23.11}  \\
Sentiment Understanding  		& 4.00  	& 4.00   			& 1.43	 	& 5.80      & 8.18  		& 12.94 	& 7.49  		& 20.09 	& 17.49 			& \textbf{20.41}    &  9.22   		&  14.43  \\ \midrule
Overall (Arithmetic MPT)		& 11.10 	& 31.11   			& 9.49 		& 39.31     & 55.25 		& 57.03 	& 60.87 		& 65.75 	& 66.07 			& \textbf{67.90} 	&  62.59   		&  67.81  \\
Overall (Harmonic MPT)  		& 0.00  	& 17.53   			& 1.92 		& 25.93     & 44.13 		& 50.30 	& 42.80 		& 58.03 	& 55.43 			& \textbf{60.47}  	&  51.87   		&  59.00  \\ \midrule
Overall (Arithmetic N-MPT)  	& 4.87  	& 15.63   			& 5.82 		& 21.46     & 29.47 		& 28.10 	& 31.36 		& 39.81 	& 35.49 			& \textbf{42.24}  	&  34.00   		&  41.04  \\
Overall (Harmonic N-MPT)  		& 0.00  	& 0.83    			& 1.91 		& 8.42      & 14.99 		& 18.30 	& 9.46  		& 24.77 	& 23.20 			& \textbf{27.28}   	&  16.67   		&  23.99   \\\midrule
Simple Accuracy          		& 21.14 	& 51.15   			& 14.54 	& 62.74   	& 69.53 		& 63.30 	& 81.07 		& 79.20 	& 78.06 			& 81.86   			&  79.56   		&  \textbf{84.26}    \\ \bottomrule
\end{tabular}
\end{table*}

\section{Additional Experimental Results}

\begin{table}
\centering
\footnotesize
\caption{Results on TDIUC-Tail for MCB model when trained on full TDIUC dataset vs when trained only on TDIUC-Tail. The un-normalized scores for each question-types and five different overall scores are shown here}
\label{tbl:results-umpt-ablation}
\begin{tabular}{lrrr}
\toprule &\multicolumn{1}{c}{\textbf{\begin{tabular}[c]{@{}c@{}}MCB  \\ TDIUC-Full\end{tabular}}} & \multicolumn{1}{c}{\textbf{\begin{tabular}[c]{@{}c@{}}MCB \\ TDIUC-Tail\end{tabular}}} \\ \midrule
Scene Recognition        		& 61.64		& 66.59  	\\
Sport Recognition        		& 71.61  	& 93.74   	\\
Color Attributes         		& 6.83	 	& 84.34   	\\
Other Attributes         		& 32.80 	& 43.37   	\\
Activity Recognition     		& 51.79 	& 74.40   	\\
Positional Reasoning     		& 25.16 	& 29.59  	\\
Object Recognition      	 	& 63.90 	& 75.89  	\\
Absurd                   		& N/A 		& N/A		\\
Utility and Affordances  		& 16.67 	& 17.59   	\\
Object Presence          		& N/A	 	& N/A  	\\
Counting                 		& 4.87  	& 29.83   	\\
Sentiment Understanding  		& 41.30  	& 50.72   	\\ \midrule
Overall (Arithmetic MPT)		& 37.66  	& 51.61  	\\
Overall (Harmonic MPT)  		& 17.51  	& 43.27  	\\ \midrule
Overall (Arithmetic N-MPT)		& 19.49  	& 34.44  	\\
Overall (Harmonic N-MPT)  		& 11.37  	& 22.32  	\\\midrule
Simple Accuracy          		& 38.55 	& 50.11  	\\ \bottomrule
\end{tabular}
\end{table}

\begin{table}
\centering
\footnotesize
\caption{Results on TDIUC-Tail for MCB model when trained on full TDIUC dataset vs when trained only on TDIUC-Tail. The normalized scores for each question-types and five different overall scores are shown here}
\label{tbl:results-nmpt-ablation}
\begin{tabular}{lrrr}
\toprule &\multicolumn{1}{c}{\textbf{\begin{tabular}[c]{@{}c@{}}MCB  \\ TDIUC-Full\end{tabular}}} & \multicolumn{1}{c}{\textbf{\begin{tabular}[c]{@{}c@{}}MCB \\ TDIUC-Tail\end{tabular}}} \\ \midrule
Scene Recognition        		& 24.86 	& 29.18  	\\
Sport Recognition        		& 54.82  	& 62.74   	\\
Color Attributes         		& 7.03  	& 84.40   	\\
Other Attributes         		& 13.04 	& 17.01   	\\
Activity Recognition     		& 45.48 	& 64.83   	\\
Positional Reasoning     		& 7.46  	& 10.99  	\\
Object Recognition      	 	& 12.55 	& 24.20  	\\
Absurd                   		& N/A 		& N/A		\\
Utility and Affordances  		& 12.37 	& 14.02   	\\
Object Presence          		& N/A	 	& N/A  	\\
Counting                 		& 4.87  	& 18.96   	\\
Sentiment Understanding  		& 12.45  	& 18.08   	\\ \midrule
Overall (Arithmetic MPT)		& 37.66  	& 51.61  	\\
Overall (Harmonic MPT)  		& 17.51  	& 43.27  	\\ \midrule
Overall (Arithmetic N-MPT)		& 19.49  	& 34.44  	\\
Overall (Harmonic N-MPT)  		& 11.37  	& 22.32  	\\\midrule
Simple Accuracy          		& 38.55 	& 50.11  	\\ \bottomrule
\end{tabular}
\end{table}

In this section, we present additional experimental results that were omitted from the main paper due to inadequate space. First, the detailed normalized scores for each of the question-types is presented in Table \ref{tbl:results}. To compute these scores, the accuracy for each unique answer is calculated separately within a question-type and averaged. Second, we present the results from the experiment in section \ref{sec:rare} in table \ref{tbl:results-umpt-ablation} (Unnormalized) and table\ref{tbl:results-nmpt-ablation} (Normalized). The results are evaluated on TDIUC-Tail, which is a subset of TDIUC that only consists of questions that have answers repeated less than 1000 times (uncommon answers). Note that the TDIUC-Tail excludes the absurd and the object presence question-types, as they do not contain any questions with uncommon answers. The algorithms are identical in both Table \ref{tbl:results-umpt-ablation} and \ref{tbl:results-nmpt-ablation} and are named as follows:
\begin{enumerate}
\item\textbf{ MCB TDIUC-Full} : MCB model trained on whole of the TDIUC dataset and evaluated on TDIUC-Tail.
\item \textbf{MCB TDIUC-Tail} : MCB model trained and evaluated on TDIUC-Tail.
\end{enumerate}

\end{appendices}

\end{document}